\newif\ifColor

\documentclass[11pt,twoside,a4paper]{book}
\usepackage[square, sort, numbers]{natbib}
\usepackage{breakcites} 
\usepackage{microtype} 

\title{Graphs as Tools to Improve Deep Learning Methods}
\author{Carlos Lassance \\ Myriam Bontonou \\ Mounia Hamidouche \\ Bastien Pasdeloup \\ Lucas Drumetz \\ Vincent Gripon}
\date{}
\usepackage{subcaption}
\expandafter\def\csname ver@subcaption.sty\endcsname{}

\usepackage[acronym, nonumberlist, shortcuts, toc]{glossaries}

\usepackage{zref-abspage}

\usepackage[vcentering,dvips]{geometry}
\usepackage{venndiagram}
\usepackage{epsfig}
\usepackage{tikz}
\usepackage{pgfplotstable}
\usepackage{makecell}
\pgfplotsset{compat=1.14}
\usepackage[T1]{fontenc}

\usepackage{amsmath}
\usepackage{amsfonts}
\usepackage{amsthm}
\usepackage{multirow}
\usepackage{colortbl}
\usepackage{booktabs}

\usepackage{bm}

\usepackage{float}

\usepackage[nottoc]{tocbibind}
\usepackage{caption}

\usepackage{fancyhdr}
\usepackage{tikz}
\usetikzlibrary{arrows}
\usetikzlibrary{shapes}

\usepackage{amssymb}
\usepackage[chapter]{algorithm}
\usepackage[noend]{algpseudocode}

\theoremstyle{definition}

\usepackage{array}
\newcolumntype{P}[1]{>{\centering\arraybackslash}p{#1}}

\newcommand{\changefont}{
    \fontsize{9}{11}\selectfont
}
\def\eqref#1{equation~\ref{#1}}

\def\1{\bm{1}}
\newcommand{\dataset}{\mathcal{D}}
\newcommand{\trainset}{\mathcal{D_{\mathrm{train}}}}

\newcommand{\validset}{\mathcal{D_{\mathrm{valid}}}}

\newcommand{\testset}{\mathcal{D_{\mathrm{test}}}}
\newcommand{\classifier}{g}
\newcommand{\featureextractor}{h} 
\newcommand{\featuremaps}{F}

\def\vf{{\bm{f}}}

\def\vlambda{{\bm{\lambda}}}

\def\vs{{\bm{s}}}

\def\vx{{\bm{x}}}
\def\vy{{\bm{y}}}
\def\vz{{\bm{z}}}

\def\evlambda{{\lambda}}

\def\evs{{s}}

\def\mA{{\bm{A}}}

\def\mD{{\bm{D}}}
\def\mE{{\bm{E}}}
\def\mF{{\bm{F}}}

\def\mH{{\bm{H}}}

\def\mL{{\bm{L}}}

\def\mS{{\bm{S}}}

\def\mX{{\bm{X}}}
\def\mY{{\bm{Y}}}
\def\mZ{{\bm{Z}}}

\def\mLambda{{\bm{\Lambda}}}

\DeclareMathAlphabet{\mathsfit}{\encodingdefault}{\sfdefault}{m}{sl}
\SetMathAlphabet{\mathsfit}{bold}{\encodingdefault}{\sfdefault}{bx}{n}

\def\features{{d}}

\def\gG{{\mathcal{G}}}

\def\adjmatrix{{\mathbf{A}}}
\def\eadjmatrix{{A}}
\def\identity{{\mathbf{I}}}

\def\sD{{\mathbb{D}}}
\def\sE{{\mathbb{E}}}
\def\sF{{\mathbb{F}}}

\def\sV{{\mathbb{V}}}

\def\sX{{\mathbb{X}}}

\def\emLambda{{\Lambda}}
\def\emAdjacency{{\mathcal{A}}}
\def\emA{{A}}

\def\emD{{D}}

\def\emH{{H}}

\def\emS{{S}}

\def\emW{{W}}

\newcommand{\R}{\mathbb{R}}

\newcommand{\sfilter}{\mathbf{\mathfrak{s}}}

\usepackage{adjustbox}

\usepackage[pdffitwindow=false,
pdfview=FitH,
pdfstartview=FitH,
pagebackref=true,
breaklinks=true,
\ifColor
colorlinks,
\fi
bookmarks=false,
plainpages=false]{hyperref}

\DeclareRobustCommand{\[}{\begin{equation}}
\DeclareRobustCommand{\]}{\end{equation}}

\usepackage[inline]{enumitem}
\newlist{inlinelist}{enumerate*}{1}
\setlist*[inlinelist,1]{label=\roman*),itemjoin={{, }},itemjoin*={{, and }}}

\usepackage{etoc}
\usepackage{titlesec}
\theoremstyle{definition}

\usepackage[utf8]{inputenc}
\usepackage{pgfplots}
\usepgfplotslibrary{groupplots,dateplot}
\usetikzlibrary{patterns,shapes.arrows}
\usepgfplotslibrary{statistics}
\pgfplotsset{compat=newest}
\usepackage{svg}

\usetikzlibrary{decorations.markings}
\tikzset{
    set arrow inside/.code={\pgfqkeys{/tikz/arrow inside}{#1}},
    set arrow inside={end/.initial=>, opt/.initial=},
    /pgf/decoration/Mark/.style={
        mark/.expanded=at position #1 with
        {
            \noexpand\arrow[\pgfkeysvalueof{/tikz/arrow inside/opt}]{\pgfkeysvalueof{/tikz/arrow inside/end}}
        }
    },
    arrow inside/.style 2 args={
        set arrow inside={#1},
        postaction={
            decorate,decoration={
                markings,Mark/.list={#2}
            }
        }
    },
}

\definecolor{imtatlantique}{rgb}{0.005,0.715,0.867}

\allowdisplaybreaks
\makeglossaries

\newacronym{GSP}{GSP}{graph signal processing}
\newacronym{DNN}{DNN}{deep neural network}
\newacronym{GFT}{GFT}{graph Fourier transform}
\newacronym{MARGIN}{MARGIN}{model analysis and reasoning using graph-based interpretability}
\newacronym{VBL}{VBL}{visual-based localization}
\newacronym{SGC}{SGC}{simplified graph convolution }
\newacronym{GKD}{GKD}{ graph Knowledge Distillation}
\newacronym{RKD-D}{RKD-D}{  Knowledge Distillation}

\begin{document}
\setlength{\parskip}{0.25 \baselineskip}
\newlength{\figwidth}
\setlength{\figwidth}{26pc}
\newlength{\notationgap}
\setlength{\notationgap}{1pc}

\frontmatter

\mainmatter
\etocsettocstyle{}{} 
\maketitle
\chapter{Graphs as Tools to Improve Deep Learning Methods}
\localtableofcontents

\section{Introduction}\label{introduction}
In recent years,  \glspl{DNN} have known an important rise in popularity. A milestone was reached in 2012, when for the first time, a \gls{DNN} called AlexNet~\cite{krizhevsky2012imagenet} won the CVPR LSRV challenge~\cite{russakovsky2015imagenet} by a large margin. Since then, outstanding performance on many important problems have been achieved with \glspl{DNN}, particularly in the fields of computer vision \cite{he2016deep} \cite{krizhevsky2012imagenet}, natural language processing \cite{vaswani2017attention} and gaming~\cite{silver2017mastering}. Other domains such as text translation~\cite{johnson2017google}, speech to text~\cite{xiong2018microsoft}, person identification and verification~\cite{hermans2017defense}, help in screening and diagnosis in medicine~\cite{burt2018deep}, object detection~\cite{zhao2019object}, user behavior study~\cite{ma2019learning} or even art restoration~\cite{gupta2019restoration} have also gained momentum in the past few years. This performance of \gls{DNN} systems in a large number of application domains is largely due to the accessibility of large training datasets and to large computational processing resources as with GPUs~\cite{hacene2019processing}.

\gls{DNN} models are thus state-of-the-art in many machine learning challenges, even if they have limitations. For example, \glspl{DNN} require a lot of training data, which might not be available in some practical applications. In addition, even though trained models seemingly perform well, if small perturbations are added to their inputs, they are prone to misclassification errors~\cite{42503}. This susceptibility can be critical in some applications such as self-driving vehicles. Furthermore, \gls{DNN} models are often viewed as ``black-boxes'' and as such their decisions are often criticized for their lack of \emph{interpretability}~\cite{taylor2006methods}. Indeed, there are a lot of domains that would benefit from explainable models, among which product-recommendation, targeted advertising, medical diagnosis\dots

There is a clear explanation to the above-mentioned shortcomings. Indeed, \glspl{DNN} were introduced as a mean to replace ``hand-crafted'' features with learned ones. This is performed thanks to the use of automatic differentiation to train a model end-to-end, meaning that only the output is explicitly constrained during the learning process. And yet \glspl{DNN} are compositional models obtained by assembling elementary functions called \emph{layers}. As such, they produce multiple intermediate representations when processing an input element. The intermediate representations are rarely explicitly constrained during training with the intention of removing any arbitrary human intervention. However, giving more latitude to the training process brings the disadvantage of harder interpretability, unwanted behaviors and more data requirements.

In recent years, authors have shown that processing multiple inputs at once can help in designing more efficient models. This is the key ingredient behind the popular \emph{batch-normalization}~\citep{ioffe2015batch} layer. Batch-normalization introduced the promising idea of constraining the intermediate representations of a batch of inputs relatively to each other. As such, there is no explicit constraint on a single representation, that is still freely optimized during the training phase. But multiple representations are constrained to exhibit certain behaviors, such as competition in the case of batch-normalization. More generally, exploiting the relations between different samples of the training set is a direction of research to address the above-mentioned problems. 

In this chapter, we review recent works that aim at using graphs as tools to improve deep learning methods. These graphs are defined considering a specific layer in a deep learning architecture. Their vertices represent distinct samples, and their edges depend on the similarity of the corresponding intermediate representations. The same principle can be deployed to define graphs at the input, or at the output of deep learning architectures. These graphs can then be leveraged using various methodologies, many of which built on top of \gls{GSP}~\citep{shuman2013emerging}.

The outline of this chapter is as follows. We first introduce the needed concepts from \gls{DNN} and \gls{GSP} frameworks in Section~\ref{background}. We then review the recent contributions aiming at using graphs to improve deep learning methods in Section~\ref{mainsection}. This Section is composed of four main parts: tools for visualizing intermediate layers in a \gls{DNN}, denoising data representations, optimizing graph objective functions and regularizing the learning process. Section~\ref{conclusion} provides some concluding remarks.

\section{Background}
\label{background}

\subsection{Deep Neural Networks for Computer Vision}\label{dnn-cv}

In this section, we introduce the fundamental aspects about \glspl{DNN} that will be of use in the remaining of the document.

A DNN is a mathematical function $f$ that associates an input $\mathbf{x}$, typically in a multidimensional array, with a corresponding output $\hat{\mathbf{y}} = f(\mathbf{x})$. It is obtained by assembling elementary functions that are called \emph{layers} in the literature. In the simplest case, this assembly consists in a simple composition of the layer functions, so that the DNN can be mathematically described as \begin{equation} f = f^{\ell_\text{max}}\circ f^{\ell_\text{max}-1} \circ \dots \circ f^1,\end{equation} where each $f^i$ is a layer function. In more complex cases, it is often the case that a DNN function can be decomposed as $f = g \circ h$, where $g$ and $h$ are possibly complex assemblies themselves.

When processing an input element $\mathbf{x}$, it is therefore possible to decompose two parts $\mathbf{z} = h(\mathbf{x})$ and $\hat{\mathbf{y}} = g(\mathbf{z})$. In such a case, we call $\mathbf{z}$ an \emph{intermediate representation} associated with $\mathbf{x}$. In practice there are multiple intermediate representations associated with an input $\mathbf{x}$ for multiple layers in the considered DNN. 

Most layers contain trainable \emph{parameters}. Their values are initialized at random and updated during a training phase. This phase usually consists in minimizing a loss function, which is a function of the discrepancy between the actual output $\hat{\mathbf{y}} = f(\mathbf{x})$ of the DNN and the expected one $\mathbf{y}$. This loss is minimized on a \emph{training set}.
In the literature, many architecture designs $f$ can be found, some of which are particularly popular. One example is the residual network (ResNet)~\cite{he2016deep}, that typically achieves good performance with a limited design complexity.

As various designs exist, it is common to rely on cross validation to choose the one that yields the best performance.  In that case, the training set is split into two parts prior to training. The first part is used as a new training set for the DNN. The second part, called the \emph{validation set} is used to pre-evaluate the ability of the trained model to generalize to new data. More precisely, at the end of the training, we obtain the final architecture with the corresponding parameters. Then, one evaluates the accuracy of the DNN on the validation set. The accuracy is the number of correctly predicted data points out of all the validation set. The chosen design is the one that generalizes better. When the error on the training set is small, whereas the error on previously unseen inputs is high, the DNN is \emph{overfitting}. To assert that the chosen DNN does not overfit and can be used to process previously unseen inputs, it is evaluated on a third dataset, called \emph{test set}.   

\subsubsection{Limitations of DNNs}

Considering a large amount of labeled data, and extensive computational power, DNNs are in many cases the golden standard of machine learning problems. However, major limitations refrain the adoption of these methods in specific domains. This chapter is built around four limitations that are all tackled using graph-based representations.

The first limitation is the \textbf{lack of interpretability} of the decisions taken by DNNs.
This is problematic for applications that aim at assisting human experts (e.g. assisted medical diagnosis). 
Since it is difficult to understand how the DNN processes information, its ability to take a good decision on new inputs is assessed via the test set. However, a DNN performing well on a test set is not guaranteed to perform well on slightly different or unexpected inputs. Various methods have been proposed to visualize the features that matter in the decision process, both in the inputs and in the learned intermediate representations~\citep{ribeiro2016lime,anirudh2017influential,szegedy2014intriguing}. To directly measure the ability of generalization, without necessarily needing a validation set, some metrics have also been proposed~\citep{gripon2018insidelook,jiang2019predicting,jiang2020fantastic,bontonou2020predicting}. In Subsection~\ref{visualize}, graph-based measures are introduced to assess the ability of generalization of DNNs. 

The second limitation is the difficulty to \textbf{train models with little data}. In the literature, learning on very few labeled examples is called \emph{few-shot learning}. More generally, we a few-label is a task that is described as $N$-way $K$-shot $U$-unlabeled, where $N$ is the number of classes to distinguish within data, $K$ the number of labeled images available per class during training, and $U$ unlabeled images per class may be available for training. In the case of a few-shot task $U$ is always equal to 0.

Many popular solutions for few-shot problems rely on \emph{knowledge transfer}~\citep{tan2018transfersurvey, zhai2019largescale}. The core idea is to exploit a DNN \nolinebreak{$f' = g' \circ h'$} trained on a distinct, yet similar and larger dataset, for the newly considered problem. More precisely, instead of training a new DNN $f$ directly on raw inputs $\mathbf{x}$, a function $g$, taking $h'(\mathbf{x})$ as its input, is trained. In such a case, $h'$ is called a \emph{feature extractor} and $g$ a \emph{classifier}, $f = g \circ h'$ is therefore the new DNN function.
The presented approach is preferred when the small dataset is very limited~\citep{arandjelovic2016netvlad, mangla2019manifold}. If more data is available, it is also common to \emph{fine tune} $h'$ as well by training $f$ for a few epochs~\citep{alex2019large}. An epoch refers to a pass of the gradient descent algorithm through the entire training set.
Subsection~\ref{denoise} introduces ways of using graphs to help denoising the representations learned by the feature extractors, in order to better handle the underlying problem. 

The third limitation lies on the fact that \textbf{the output of the DNN is most of the time independent of the distribution of the input and of the initialization
of the network parameters}, which can slow down and complicate the
training process.
To understand this limitation, consider the cross-entropy loss function, the most popular loss function for classification problems in computer vision. In this case, the dimension of the output vectors has to be equal to the number of classes, preventing an easy adaptation to the introduction of new classes. In scenarios where the number of classes is large, this also causes the last layer of the network to contain a lot of parameters.  Moreover, the cross-entropy loss considers that it is possible to completely separate the classes but does not take into account a possible hierarchy or similarity between classes. To overcome these drawbacks, one alternative to this is to optimize the embeddings directly, without forcing arbitrary separation such as the case of the siamese~\citep{koch2015siamese}, triplet~\citep{hermans2017defense} and smoothness losses~\citep{bontonou2019smoothness}. This limitation is addressed in Subsection~\ref{losses}.

The last considered limitation is the \textbf{lack of robustness} that occurs when DNN architectures are susceptible to deviations of their inputs and thus perform poorly~\cite{szegedy2014intriguing,goodfellow2014adversarial}. The lack of robustness may cause errors in decisions in various practical settings, in which data is prone to noise or manipulations. When discussing robustness of DNNs, it is common to separate two sources of deviations. The first source of deviation is the one that is not performed intentionally to fool the DNN~\citep{hendrycks2019robustness}\footnote{such deviations are called \emph{corruptions} in this chapter.}. Examples of such deviations range from defects in the data capture (e.g. Gaussian noise) to different luminosity conditions (e.g. brightness and contrast). The second deviation is the one that is created explicitly to fool the network, also called adversarial attacks. In this case, a metric is used to measure the distance between the original image and the adversarial one~\citep{goodfellow2014adversarial}. A successful adversary is the one that fools the DNN, without trespassing a perceptibility threshold on the metric (e.g. the adversary does not change any individual pixel color of more than 8/255~\citep{madry2018towards}). This limitation is addressed in Subsection~\ref{regularizers}.

\subsection{Graph Signal Processing}\label{gsp}

Graph Signal Processing (GSP) offers a convenient framework for studying data while taking into account the complex domain on which they are defined. This section introduces the main tools that are of use in GSP. In the remainder of this chapter, we will use them to express desirable constraints on DNNs.

\subsubsection{Graphs and Graph Signals}\label{gsp:definitions}

Consider a dataset comprising $n$ samples, and suppose that there exist specific relations between them. For example, samples can be individuals in a social network and relations are given by their friendship connections. A concise and convenient way to model samples and their relations is to use the formalism of graphs.

A \emph{graph} $\gG$ is a tuple of sets $\langle \sV , \sE \rangle$, where $\sV$ is composed of \emph{vertices} indexed from 1 to $n$: $\sV = \{v_1,\dots,v_n\}$, and $\sE$ is composed of pairs of vertices of the form $(v_i,v_j)$ called \emph{edges}. These edges are unordered, such that $(v_i, v_j)$ refers to the same pair as $(v_j,v_i)$.

It is common to represent the set $\sE$ using an edge-indicator symmetric \emph{adjacency matrix} $\adjmatrix \in \R^{|\sV|\times |\sV|}$. When relations are weighted, this matrix can take values other than 0s and 1s.

In a graph, some vertices might have more importance than others. A common way to measure importance of vertices is to define their \emph{degree} (or \emph{strength} when graphs are weighted). The degree of a vertex $v$ is simply obtained by summing the weights of all edges $v$ is part of. These quantities can be assembled in the diagonal \emph{degree matrix} $\mD$:
\begin{equation}
    \emD_{i,j} = \left\{ \begin{array}{cl}\displaystyle{\sum_{j' \in \sV}{\eadjmatrix_{i,j'}}} & \text{if } i = j\\ 0 & \text{otherwise.}\end{array}\right.\;.
\end{equation} 

Now let us suppose that each of the $n$ samples is associated with a vector of dimension $\features$. This can be represented as a $n \times \features$ matrix. Columns of this matrix are called graph signals. For example, the age of each individual in a social network can be viewed as a graph signal.

As a natural representation of complex data structures, graphs and graph signals are ubiquitous, in particular in the field of machine learning. In the following paragraphs we introduce operations on graphs that are going to be useful in the rest of this chapter.

\subsubsection{Graph Fourier Transform (GFT)}

In the domain of signal processing, \emph{frequencies} are one of the most important concepts to analyze a signal and a starting point to introduce many useful tools such as filtering. The frequency of a signal can be simplified as its rate of change, or in other words how it varies from one sample to another. Generally speaking, any signal can be decomposed into a (possibly continuous) sum of sines/cosines of various frequencies, by performing the celebrated Fourier transform. This decomposition can be understood as expressing the signal in the frequency domain, providing a dual representation of the original signal in the time domain. In other words, the Fourier transform of a signal is a simple change of basis, that transports a signal to a convenient domain for many signal processing operations.

Graph signal processing arose as a graph-centric generalization of the classical Fourier analysis \cite{shuman2013emerging}. In this framework, a ring or a path-shaped graph is used to model the time domain, in which each vertex models an instant in time where the intensity of the signal is sampled. One therefore manipulates two separate objects: a graph $\gG$ modeling time -- the support of information -- and a graph signal $\vs$ -- the sampled information --. In a path graph, each vertex is connected with at most one subsequent vertex (in this case the next sample in the time domain) and at most one previous vertex (in this case the previous sample in the time domain). This means that in a path graph there exist two vertices with only one edge, called ``first'' and ``last''. A regular ring graph is then defined as a path graph where the ``first'' vertex is connected to the ``last'' (one-dimensional lattice), forming a cycle with periodic boundary conditions, that is very much a natural subject for Fourier analysis. 

The core observation of GSP is that Fourier transform is closely tied to the domain over which signals evolve, and not to the signals themselves. More precisely, consider a graph whose adjacency matrix only contains nonnegative values. Its combinatorial Laplacian $\mL$ is defined as:
\begin{equation}
    \mL = \mD - \adjmatrix \; .
\end{equation} 
As the Laplacian matrix is both real and symmetric it can be eigendecomposed into:
\begin{equation}\label{gsp:laplacian_spectral}
    \mL = \mF \mLambda \mF^\top \; ,
\end{equation} 
where the columns of $\mF$ are the eigenvectors of $\mL$ and the diagonal coefficients of $\mLambda$ are the eigenvalues in increasing order of magnitude.

If we consider the regular ring graph mentioned above that has a periodic property to model time as a support of information, one can show that the eigenvectors in $\mF$ are described as sine/cosine functions (Fourier series), with increasing frequency as associated eigenvalues increase. The Fourier transform of a graph signal $\vs$ is thus simply a change of basis defined by $\mF$, and can be written as:
\begin{equation}
    \hat{\vs} = \mF^\top \vs \; .
\end{equation} 

This projection of $\vs$ in the eigenbasis $\mF$ of $\mL$ is what we call the \emph{Graph Fourier Transform (GFT)} of $\vs$. Note that the inverse GFT can be similarly defined as:
\begin{equation}
    \vs  = \mF \hat{\vs} \; .
\end{equation} 

GFT obviously depends on the graph that is used to model the signals under study. As a consequence, the choice of the underlying graph is of paramount importance. A path (or ring) graph provides a comprehensive domain to study (periodic) time signals, but it is often not clear which graph should be used for more complex signals. However, it still holds that eigenvectors $\mF$ of the Laplacian matrix of a graph characterize the variations over that graph, with increased local variations as the associated eigenvalues increase. The interested reader can refer to \cite{shuman2013emerging} and \cite{ortega2018graph} for illustrative examples.

Exploiting this property, authors have defined many tools generalizing signal processing operations on temporal signals, such as convolution, filtering, translation or modulation~\cite{shuman2013emerging}. We present in the next sections some of the uses of the GFT relevant to the works introduced in this chapter.  

\textbf{Smoothness of graph signals}\label{gsp:smoothness}
A measure that will be at the core of many of the tools we introduce in the next section is the \emph{graph smoothness}. Let $\gG$ be a graph of adjacency matrix $\adjmatrix$, and let $\mL$ be its combinatorial Laplacian, with eigenvalues $\mLambda$. The smoothness $\sigma(\vs)$ of a graph signal $\vs$ supported on $\gG$ is defined as:
\begin{equation}
    \sigma(\vs) = \vs^\top \mL \vs = \sum_{i=1}^{|\sV|} \emLambda_{i,i} \hat{\vs}_{i}^2 = \sum_{i=1}^{|\sV|}\sum_{j=1}^{|\sV|}\eadjmatrix_{i,j} (\vs_i - \vs_j)^2 \;.
    \label{eq:smoothness}
\end{equation}
Lower values of the smoothness metric are said to be very smooth with respect to the graph, and higher values of the smoothness metric are said to be very rough (or unsmooth) with respect to the graph.
Equation (\ref{eq:smoothness}) shows that a smooth signal is a signal aligned with the first eigenvectors of the Laplacian. In addition, we observe that the smoothness is strongly related to the rate of change of the signal values from one vertex to its neighbors.

Considering the smoothness as the rate of change in a signal $\vs$ across a graph, it is a very useful abstraction, even more when $\vs$ is binary. In this case, the smoothness can be simplified as the sum of the weights between neighboring nodes with different values in $\vs$. 

Consider an illustrative case where vertices represent samples in a classification dataset. An interesting signal is the label indicator vector of a chosen class, which is a binary vector that contains a 1 at coordinate $i$ if $i$ is of the considered class and 0 otherwise. In such a case, the smoothness boils down to the sum of the weights of edges connecting pairs where one element is of the considered class and the other is not. Therefore, the only graphs that would nullify smoothness of this signal are those that contain no edge between samples of the class and samples of a distinct class.

\textbf{Graph filters}\label{gsp:graph_filter} In classical discrete signal processing, filters are tools meant to act on signals by removing specific ranges of frequencies. In the following, we develop the equivalent concept of graph filters for graph signals in GSP.

Here, we introduce three possible representations of these filters. For each representation, we provide its advantages/drawbacks and detail some of their applications. We refer the reader to~\citep{hammond2011wavelets, tremblay2018design} for a more in depth discussion on graph filters.

\textbf{1) Design of filters in the spectral domain}
The simplest and most general way to define a filter is to describe its response for each frequency. In the case of graph signals, frequencies are defined by the diagonal values of the eigenvalue matrix $\mLambda$. For notation simplicity we consider the frequency vector $\vlambda$ where $\evlambda_i = \emLambda_{i,i}$, sorted in increasing order of magnitude. 

We define a filter on a graph $\gG$ using a diagonal matrix $\mH_{\gG} \in \R^{|\sV|\times |\sV|}$, where we call each element $\emH_{i,i}$ the response of the filter to frequency $\lambda_i$. The filter can then be convoluted with a graph signal $\vs$ as in classical signal processing, by performing a multiplication in the spectral domain \cite{shuman2013emerging}:

\begin{equation}
\tilde{\vs} = \mF \mH_{\gG} \mF^\top \vs \;.
\end{equation}
  
Defining filters directly in the spectral domain generates filters that are graph-specific. Indeed, as the frequencies $\evlambda_i$ are discrete, the same filter $\mH$ will be applied to very different frequencies for two distinct graphs $\gG$ and $\gG'$. Therefore this type of filter tends to be mostly used to remove the lowest or highest frequencies of the graph, without considering their ``true'' value. Another drawback is that it is unlikely that this type of filter may be represented with a low order polynomial filter which impacts the complexity (i.e. the possibility of scaling to larger graphs) of applying the filter.

\textbf{2) Design of filters using their spectral response}

\begin{figure}[ht]
\centering
\begin{minipage}{.4\textwidth}
\begin{equation*}
  h(\evlambda_i)=\begin{cases} 1 & \mbox{if }\frac{\evlambda_i}{\lambda_{|\sV|}}\leq \frac{\tau}{2},\\
              \cos\left(\frac{\pi}{2}\frac{\log\left(\frac{\evlambda_i}{\tau \lambda_{|\sV|}}\right)}{\log(2)}\right) & \mbox{if }\frac{\tau}{2}<\frac{\evlambda_i}{\lambda_{|\sV|}}\leq\tau,\\
              0 & \mbox{if }\frac{\evlambda_i}{\lambda_{|\sV|}}>\tau. \end{cases}
\end{equation*}
\end{minipage}
\hfill
\begin{minipage}{.5\textwidth}
%
  \begin{tikzpicture}
    \begin{axis}[
      xlabel=$\text{Frequency } \lambda_i$,
      ylabel=$\text{Filter response } h(\lambda_i)$,xmin=0,xmax=1]
        \addplot[color=red,domain=-2:-1] {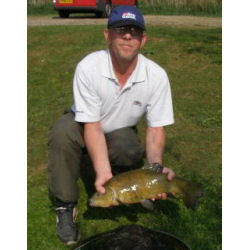};
        \addlegendentry {$\tau=0.2$}
        \addplot[color=blue,domain=-2:-1] {1};
        \addlegendentry {$\tau=0.5$}
        \addplot[color=black,domain=-2:-1] {1};
        \addlegendentry {$\tau=0.7$}

        \addplot[color=red,domain=0:0.1] {1};
        \addplot[color=red,domain=0.1:0.2] {cos( deg( (pi/2) * ln(2*x/0.2) / (ln(2)) ) ) };
        \addplot[color=red,domain=0.2:1.0] {0.0};
        \addplot[color=blue,domain=0:0.25] {1};
        \addplot[color=blue,domain=0.25:0.5] {cos( deg( (pi/2) * ln(2*x/0.5) / (ln(2)) ) ) };
        \addplot[color=blue,domain=0.5:1.0] {0.0};
        \addplot[color=black,domain=0:0.35] {1};
        \addplot[color=black,domain=0.35:0.7] {cos( deg( (pi/2) * ln(2*x/0.7) / (ln(2)) ) ) };
        \addplot[color=black,domain=0.7:1.0] {0.0};
    \end{axis}
  \end{tikzpicture}
  %

\end{minipage}
\caption{Equation (left) and depiction (right) of the Simoncelli filter spectral response for various values of $\tau$.}
\label{gsp:fig-simoncelli}
\end{figure}
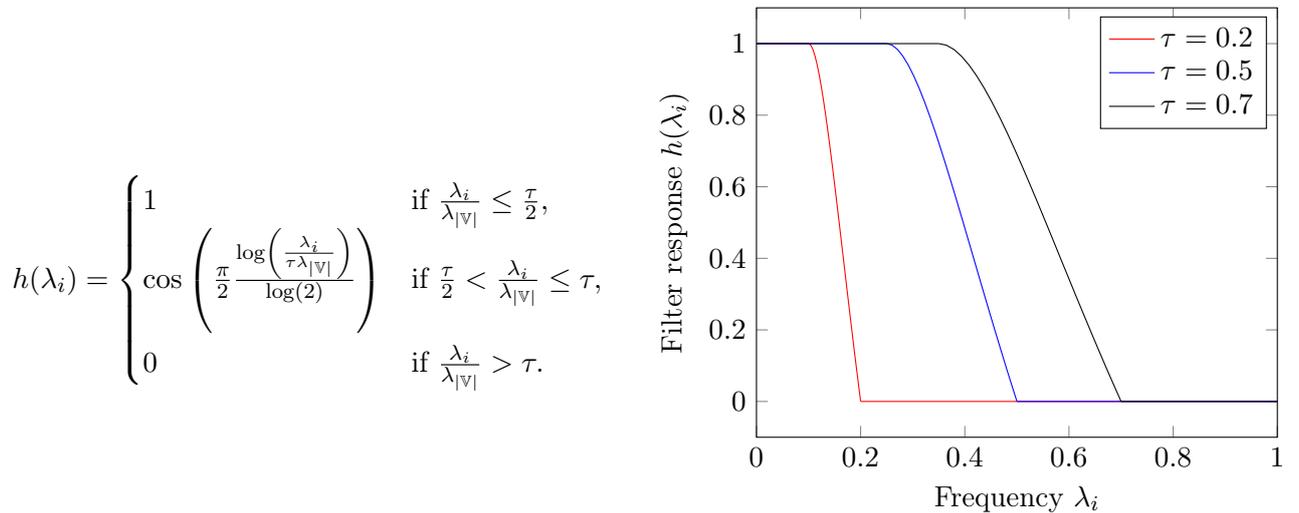

Another straightforward way to define a filter is by its spectral response, i.e. a function of the frequency. In this way the filter becomes less graph-dependent and more general. One such design is the Simoncelli filter, depicted in Figure~\ref{gsp:fig-simoncelli}, where $\tau\in [0,1]$ is a user-defined threshold and $\lambda_i$ the $i$-th Laplacian eigenvalue.

Defining a filter by its spectral response allows for more universal filters (i.e. filters that do not heavily depend on the graph support) and also to easily represent the filter by a low order polynomial function. In this work, we use the PyGSP~\citep{pygsp} toolbox to implement this type of graph filters, which uses the Chebyshev polynomial approximation in order to apply the filters.

\textbf{3) Design of filters using diffusion operators}

It is also possible to define a graph filter using a diffusion operator $\mS \in \R^{|\sV|\times |\sV|}$, sometimes also called \emph{graph shift operator}. A diffusion operator on a graph $\gG$ is a matrix $\mS$ describing the same edge set as the adjacency matrix $\mA$ of $\gG$, i.e. such that $\emS_{i,j} = \emA_{i,j}$ where $\emA_{i,j} = 0$. It allows to define the notion of information flow over a graph, as it represents one elementary discrete step on how the information propagates (shifts) from one node to its neighbors. In classical signal processing -- where the graph structure can be modeled as a directed cycle graph -- the information flow is unidirectional, and goes from each node to its only neighbor. In a more complicated graph structure, the information flow is neither restricted to unidirectional structure nor to a limited number of physical neighbors but depends on the graph adjacency matrix.
When information flows to a vertex from all the vertices it is neighbor of, incoming signal is combined through a weighted sum as follows \cite{sandryhaila2014discrete}:
\begin{equation}
  \tilde{\vs} = \mS \vs \;.
\end{equation} 
Note that by being applied with just a simple matrix multiplication, this type of filter is easily integrated in deep learning scenarios, where matrix multiplication is prevalent. Indeed, most of the recent developments in graph convolutional layers use this design.

\subsubsection{Example of application of GSP}

\begin{figure}[ht]
  \begin{center}
    \begin{subfigure}[ht]{.3\linewidth}
      \centering
      \begin{adjustbox}{max width=\linewidth}
        %
  \begin{tikzpicture}
  \begin{axis}[
    xlabel=$v$,xticklabels=none,
    ylabel=$\vs$,ymin=-3,ymax=3,ycomb]
  \addplot[mark=*] table {gsp/signals/sample_vs};
    \end{axis}
\end{tikzpicture}%

      \end{adjustbox}
      \caption{$\sigma(\vs) = 0.132$}
    \end{subfigure}
  \begin{subfigure}[ht]{.3\linewidth}
      \centering
      \begin{adjustbox}{max width=\linewidth}
        %
  \begin{tikzpicture}
    \begin{axis}[
        xlabel=$v$,xticklabels=none,
        ylabel=$\breve{\vs}$,ymin=-3,ymax=3,ycomb]
      \addplot[mark=*] table {gsp/signals/sample_tildevs};
    \end{axis}
\end{tikzpicture}%

      \end{adjustbox}
      \caption{$\sigma(\breve{\vs}) = 1270.902$}
    \end{subfigure}        
  \begin{subfigure}[ht]{.3\linewidth}
    \centering
    \begin{adjustbox}{max width=\linewidth}
      %
  \begin{tikzpicture}
    \begin{axis}[
        xlabel=$v$,xticklabels=none,
        ylabel=$\sfilter$,ymin=-3,ymax=3,ycomb]
      \addplot table {gsp/signals/retrieved_signal_gft};
    \end{axis}
\end{tikzpicture}%

    \end{adjustbox}
    \caption{$\sigma(\tilde{\vs}) = 0.144$}
  \end{subfigure}        
\caption{Depiction of a sampled signal $\vs$, its noisy version $\breve{\vs}$ and its filtered version $\tilde{\vs}$, with their respective smoothness values.}
  \label{gsp:fig-smoothness-graph}
  \end{center}
\end{figure}
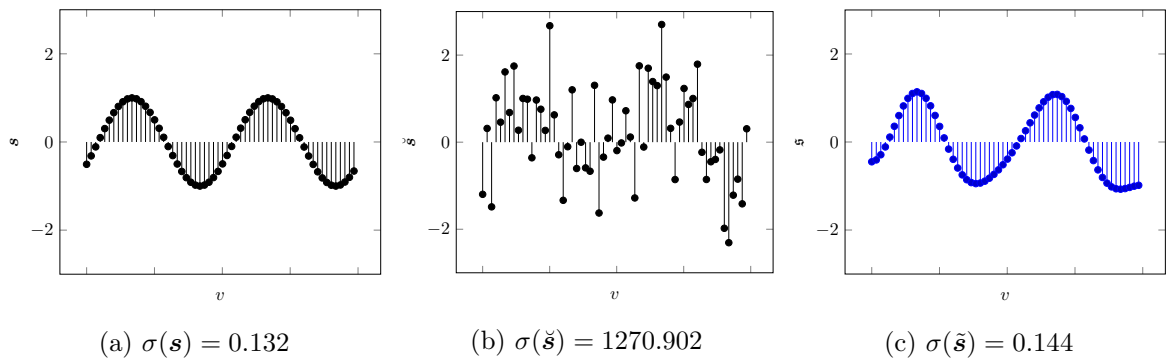

Here is an illustrative example showing the usefulness of the notion of graph filters and smoothness.  
Figure~\ref{gsp:fig-smoothness-graph} depicts a graph signal on a path graph, a noisy version of it, and a low-pass filtered version of the noisy graph signal. Note that both the original graph signal and the filtered version are smoother than the noisy version of the signal. Thus, the smoothness of a graph signal can also be used to detect the presence of noise in the signal.

\begin{figure}[ht]
  \begin{center}
      \begin{subfigure}[ht]{.3\linewidth}
        \includegraphics[width=\linewidth]{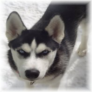}
        \caption{$\sigma \equiv 1211 \times 10^{4}$}
      \end{subfigure}
      \begin{subfigure}[ht]{.3\linewidth}
        \includegraphics[width=\linewidth]{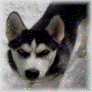}
        \caption{$\sigma \equiv 2043 \times 10^{4} $}
      \end{subfigure}
      \begin{subfigure}[ht]{.3\linewidth}
        \includegraphics[width=\linewidth]{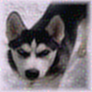}
        \caption{$\sigma \equiv 1217 \times 10^{4}$}
      \end{subfigure}
      \caption{Depiction of a dog (left), a noisy realization of the image (center) and the graph filtered image (right).}
      \label{gsp:fig-denoising}
  \end{center}
\end{figure}

In addition, here is an example of a picture, seen as a signal of dimension 3 (RGB) over a grid graph. Using a low-pass graph filter, it is possible to exploit the structure of the pixels to remove noise from the image as shown in Figure~\ref{gsp:fig-denoising}. By removing the high frequencies of the graph signal, the resultant image is more in line with the original image. Note also that the smoothness value of the recovered graph signal is the same as the original image, while the images are very different. Indeed as the smoothness of a graph signal is a global measure and not a localized one it is easy to see that multiple image configurations are possible for the same value of smoothness. 

\section{Graphs as Tools to Improve Deep Learning Methods}
\label{mainsection}

Now that we have introduced DNNs, some of their limitations and presented the framework of GSP, we delve in more details into how to improve deep learning methods using graphs.

\subsection{Graphs to Interpret Intermediate Representations}\label{visualize}

\subsubsection{Using smoothness to measure the state of DNNs} \label{visualize:smoothness}

Throughout the training process of a DNN, an input vector is transformed into multiple intermediate representations of various dimensions and shapes. Understanding how these intermediate representations evolve during the training process is a major open question.

In particular, it is difficult to know whether a trained architecture is able to solve the problem it was trained for or not. To address this limitation, cross validation is often performed to measure the state of a DNN: underfitted, overfitted or correctly optimized. However, in many practical cases, it is problematic to rely on cross validation. Indeed, the training set is reduced in size, requiring to tune parameters with a limited number of examples. Also, the model generalization abilities are assessed using a fraction of the training set, giving a possibly biased judgement on their real performance on unseen data. 

A more principled approach was proposed in~\citep{gripon2018insidelook} where the authors propose to use GSP to monitor the training of DNNs.
Their method is based on the notion of smoothness of graph signals. Namely, a graph $\gG$ is constructed at different layers of the DNN. Its vertices are the intermediate representations of the training samples and its edges are the similarities between these samples. The graph signal $\vs^c \in \R^{|\sV|}$ indicates whether the vertices belong to the class $c$. More precisely, given a vertex $i$, $\vs^c_i = 1$ if the class of the $i$-th sample is $c$ and $0$ otherwise. 

Let $M$ be the number of examples in each class $c$, $C$ be the number of classes and $\mL$ the Laplacian of $\gG$. For a given graph signal $\vs^c$, the label smoothness is then defined as:

\begin{equation*}
   Smooth = \dfrac{\sum_{c=1}^{C} \vs_c^\top \mL \vs_c}{M^2 C(C-1)}.
\end{equation*}

The term $M^2(C-1)$ is a normalization factor. The label smoothness can be rewritten as the sum of the similarities between samples of distinct classes. Thus, the label smoothness is 0 if and only if the similarity between samples of distinct classes is 0.

The authors of~\citep{gripon2018insidelook} were able to show that the label smoothness can clearly differentiate different states of trained DNNs. 
More precisely, they observed that when the considered architecture is correctly optimized,  the values of label smoothness on the last layers of the trained architecture are similar, whereas in the other conditions, they are separated by important gaps.

They also observed that during training, the label smoothness continues to evolve when the training accuracy stabilizes around 100\%. Indeed, this motivates the idea that even if the training accuracy is maximal, the intermediate representations are still changing. Therefore, the label smoothness could be used as a measure to stop or continue the DNN training when the training accuracy has reached its maximal value.

\subsubsection{Detecting the influential instances}\label{visualize:influential}
To better interpret a DNN, one can look at many different variables, e.g. the training examples that influence the most the predictions or the most influential pixels inside an image. In this subsection, a method called \emph{model analysis and reasoning using graph-based interpretability} (MARGIN)~\cite{anirudh2017influential} is presented. In this method, the studied influential variables are called \emph{influential instances}. The method is divided into four steps: \begin{enumerate}
    \item \emph{Influential instances}: First, the domain on which the analysis is performed is described. It can be the set of training samples, where each instance is an image, or even a single image, where a group of neighboring pixels is an instance.  
    \item \emph{Graph construction}: A graph $\gG$ modeling the relationships between the instances is constructed. The vertices are the instances and the edges are the similarities between the instances. The similarities can be computed from the raw instances or, when applicable, from intermediate representations of the instances within the DNN.
    \item \emph{Graph signal}: The graph signal represents the source of variations on which the influence of each instance is studied. For example, considering that an instance is a training example, the graph signal could be a measure indicating how much the neighbors of the example in $\gG$ disagree with its label.
    \item \emph{Sample-wise influence estimation}: MARGIN aims at finding the vertices of the graph $\gG$ that characterize the most the variations of the signal with respect to the structure of the graph $\gG$. Using the formalism of GSP, a high-pass filter is applied over the graph signal. The influence of each vertex is proportional to the amplitude of the filtered signal.
\end{enumerate}

In~\cite{anirudh2017margin}, the MARGIN protocol is evaluated on three tasks: showing which parts of an image influence the most the prediction made by a DNN, pointing out to the hardest training examples to classify and detecting adversarial attacks. 
Experiments are performed using a common dataset of image classification and the Alexnet~\citep{krizhevsky2012imagenet} DNN architecture.

MARGIN enables to obtain sparse saliency maps highlighting the importance of some group of pixels in the classification decision. MARGIN also enables to find hard examples. In a two-way classification task, trying to distinguish tabby cat and great dane, the images with the highest influence values (harder to classify) are the ones where the animal face is not visible. MARGIN is also useful to detect adversarial examples. An experiment shows that MARGIN induces a very sharp difference between the distribution of the adversarial examples and the one of the training examples, with only a small overlap in the distributions. These overlaps have been investigated and it was shown that they correspond to samples that are similar to the confusing examples from the previous experiment.

\subsection{Graphs to Denoise Intermediate Representations}\label{denoise}

In this section, we introduce a few examples of methods based on GSP and aiming at ``denoising'' the features extracted from DNNs. 

\subsubsection{Graph filtering for localizing images}

Visual-based localization (VBL) is the problem of retrieving the location and orientation (pose) of the camera which generated a given query image. For example, from one photograph of a place, the goal is to retrieve its GPS coordinates.
Many different photographs can be taken from one place, with different angles of view, various weathers, different people on them. Thus, the VBL problem becomes challenging because of the difficulty to learn a representation which is resilient this amount of appearance variations.

Some researchers~\citep{brahmbhatt2018geometry,kendall2015posenet} proposed to directly train a DNN to map images to GPS positions~\citep{brahmbhatt2018geometry,kendall2015posenet}. However, such method comes with serious limitations. For example, they are unable to generalize to GPS positions unseen during training. Also, appending new GPS positions to the dataset requires retraining the whole DNN. Furthermore, small differences between the training images and a new image can cause significant localization errors.

In this subsection, another method addressing both limitations is presented~\citep{lassance2019improved}. It is based on Graph Signal Processing, and aims at improving the performance of VBL by incorporating additional available information. The additional information acts on the representations of the data, making them smoother on a graph designed using all available information. The consequence is a boost
in localization performance.

The method requires both a support set of images from a predefined set of places and the access to a DNN trained to map images to representations that are more resilient to appearance changes~\citep{arandjelovic2016netvlad}. The method does not require additional learning, allowing it to be possibly executed on a resource constrained system. Interestingly, it can be extended to new places without needing to retrain as well. 

In details, the proposed method consists in denoising the representations exploiting a low-pass graph filter~\citep{arandjelovic2016netvlad,radenovic2019finetune}. In the graph $\gG$, each vertex is associated with an image. The edges model relations between images and are derived from the additional source of information (e.g. duration between two consecutive photographs, GPS positions, similarities of the representations\dots). In the following, the inference of the graph $\gG$ and the graph filter are defined more precisely.

The graph $\gG$ is described by its adjacency matrix $\mA$ built from three different sources of information:
\begin{equation}
  \adjmatrix = \adjmatrix_{\texttt{dist}} + \adjmatrix_{\texttt{seq}} + \adjmatrix_{\texttt{sim}},
\end{equation}
where $\adjmatrix_{\texttt{dist}}$ represents the distance measured by the GPS coordinates between two vertices, $\adjmatrix_{\texttt{seq}}$ represents the delay in time acquisition between two images (acquired as frames in videos), $\adjmatrix_{\texttt{sim}}$ represents the cosine similarity between the representations of the two images.

The matrix $\adjmatrix_{\texttt{dist}}$ is parametrized by a scalar $\gamma$ and a threshold parameter $\texttt{dist}_\text{max}$ cutting edges between distant vertices:
\begin{equation}
\adjmatrix_{\texttt{dist}}[i,j] = \left\{ \begin{array}{ll} e^{-\gamma \texttt{dist}_{i,j}} & \text{if } \texttt{dist}_{i,j} < \texttt{dist}_\text{max}\\ 0 & \text{otherwise}\end{array}\right..
\end{equation}

To exploit the information of time acquisition of frames, the function $seq(k, i, j)$, returns 1 if the frame distance between $i$ and $j$ is exactly $k$ and 0 otherwise. The matrix $\adjmatrix_{\texttt{seq}}$ is parametrized by scalars $\beta_k$ and $k_{max}$:

\begin{equation}
  \adjmatrix_{\texttt{seq}}[i, j] = \sum_{k=1}^{k_{max}} \beta_{k} \texttt{seq}(k,i, j).
\end{equation}

The matrix $\adjmatrix_{\texttt{sim}}$ depends on the cosine similarity $sim_\text{cos}$ between the representations of the images. It is parametrized by a scalar $\alpha$, which controls the importance of the similarity with respect to the two other sources of information. The cosine similarity between two images is only considered if their GPS positions are close or if the photographs have been taken successively:

\begin{equation}
\adjmatrix_{\texttt{sim}}[i, j] = \left\{ \begin{array}{ll} \alpha sim_\text{cos}(i, j) & \text{if } \adjmatrix_{\texttt{dist}}[i, j] > 0 \\ & \text{or } \adjmatrix_{\texttt{seq}}[i, j] > 0,\\ 0 & \text{otherwise.}\end{array}\right.
\end{equation}

Given the signal $\vs$, the normalized Laplacian matrix $\mL_\text{norm} = \mD^{-\frac{1}{2}}\mL\mD^{-\frac{1}{2}}$ and two hyperparameters $a$ and $m$, the filtered signal is defined by the equation:
\begin{equation}
  \tilde{\vs} = \left(\identity - a\mL \right)^m\vs.  
\end{equation} 
Note that when $m=0$ no filtering is performed.

The authors stressed their method using two datasets collected in Australian cities (Adelaide and Sidney). The datasets are composed of videos taken by vehicles while they were moving around the cities. The goal is to learn to retrieve the GPS position of each frame. 

The experiments show that on the Adelaide dataset, denoising increases performance, even when applied only on the test set, and as expected, using graph filters on both test and support gives the best results. Second, on the Sydney dataset, while using the parameters optimized for the Adelaide dataset, the graph filter gets better performance in both median distance and accuracy. This shows that it is not needed to re-estimate parameters for a new dataset. 

\subsubsection{Manifold denoising for few-label classification}

In classification, training a DNN requires a huge number of labeled data samples.
When a DNN is trained on a small set of samples,
it is likely to learn many irrelevant details of the data. It will not infer well the class of new samples coming to the classifier. This is a well known problem called \emph{overfitting} (see Section~\ref{dnn-cv}). To handle overfitting, one can take advantage of the fact that DNNs are really good at extracting relevant information from data. Indeed, as a reminder a DNN can often be split into two functions $f = \classifier \circ \featureextractor$.
The first part $\featureextractor$ learns to represent the data samples in a space where a simple classifier $\classifier$ can separate the data samples according to their classes. For example, when dealing with images, it is common to choose $\featureextractor$ as a convolutional neural network and $\classifier$ as a logistic regression.

The features of the data samples at the output space of the function $h$ are often assumed to be only in a smooth subspace (manifold) of the entire output space. Sometimes, because of noise within the representations, features may appear outside this manifold. Removing the noise from the representations could help in making the features closer to their underlying structure, and thus, improve the performance.

The denoising method presented in this subsection is used in a few-label setting in which few labeled samples $\mX_S$ and many unlabeled samples $\mX_U$ are given.
Consider a DNN, trained on a generic dataset, used to extract some features for a new few-shot classification task. A classifier will be trained on top of the extracted features to distinguish the new classes. However, the features may not be completely relevant for the new classification task. One may act directly on the training of the DNN using self-supervision, manifold-mixup or even training on a variety of side tasks~\cite{mangla2020charting, milbich2020diva}.
However, here, it is assumed that the features of all data samples are already given. In this case, one can exploit the additional information given by the representations of the unlabeled samples (more representative feature space). That is the case of a method based on GSP (Transfer+SGC~\citep{hu2020exploiting}) that adds a simplified graph convolution (SGC)~\citep{wu2019simplifying} on top of the extracted features. Note that SGC can be seen as a graph filter (previously defined in Section~\ref{gsp}).

The authors use the notion of cosine similarity between the representations of two data samples in the feature space. Note that in the experiments, the features are always extracted after a ReLU function, so that they contain non-negative values. Thus, the output of the cosine similarity ranges from $0$ to $1$.

The idea is to build a $k$-nearest neighbor similarity graph $\gG = \langle \sV, \sE \rangle$, where
each node represents a data sample. The connections between the nodes are weighted by the cosine similarity of their associated samples.

After removing self-loops, only the $k$ most similar neighbors are kept for each vertex. Then, the resulting adjacency matrix $\adjmatrix$ is normalized using the inverse square root of the degree matrix: $\mE = \mD^{-1/2}\adjmatrix\mD^{-1/2}$. 

Given $\mF$ the matrix containing the features of all data samples and $\identity$ the identity matrix, the new features are obtained by propagating the extracted features as follows:
\begin{equation}
    \mathbf{\mF}_{\text{diffused}} = (\alpha\identity + \mE)^m \mathbf{F}\;,
\end{equation}
where $\alpha$, $m$ and $k$ are hyperparameters. On top of the diffused features of the labeled samples, a logistic regression is trained to distinguish the labels.

The authors of~\citep{hu2020exploiting} observed that their method enabled a significant gain in the 1-shot case and a smaller one in $5$-shot with respect to existing alternatives. As a matter of fact, the added information coming from the unlabelled samples is relatively more important in the 1-shot case than it is for the 5-shot case.

\subsubsection{Manifold denoising for semi-supervised classification}

Here, a different scenario is considered in which a DNN $f = \classifier \circ \featureextractor$ is learned from scratch. It takes into input an image and outputs its class without overfitting on the training samples $\mX_S$. However, in many cases, the training data are still not representative of all the possible data in the real world. In a semi-supervised setting, one solution to better fit the real data distribution is to use the unlabeled samples $\mX_U$ as additional training data for the DNN. In the following paragraphs, an approach described in~\cite{iscen2019label} will be presented.  This approach exploits the so-called ``manifold assumption'': similar examples in the input space of the classifier $g$ should have the same labels. This assumption is applied to label the unlabelled examples, so that they can be used as additional training data.

The method consists in merging the label propagation method, which is a classical semi-supervised learning method, with a DNN.
First, a DNN is trained on the labeled examples to minimize a classification loss defined on the labeled images as:
\begin{equation}
    L_s(X_S, Y_S, \theta) = \sum_{i=1}^l l_s(\featureextractor(\vx_i, y_i)\;,
\end{equation}
where $l_s$ is the cross-entropy loss.

A nearest-neighbor similarity graph is built at the input space of the classifier $\classifier$. More precisely, each node of the graph corresponds to an image (labeled or unlabeled). The adjacency matrix $\adjmatrix$ of the graph is defined so that $\adjmatrix_{i,j}$ measures the similarity between $\vf_i=\featureextractor(\vx_i)$ and $\vf_j=\featureextractor(\vx_j)$. After removing self-loops ($\adjmatrix_{i,i}=0$), only the $k$-th largest values on each row of $\adjmatrix$ are kept. To make $\adjmatrix$ symmetric, its transposed version is added to it. As in the previous paragraphs, a normalized version of the adjacency matrix is used.

The core idea of the approach is to propagate the known labels on the similarity graph to define pseudo-labels for the unlabeled images. Indeed, a label matrix $\mY \in \R^{n\times c}$ is defined as:
\begin{equation}
    \mY_{i,j} = \left\{
    \begin{array}{ll}
    1 & \mbox{if } 1\leq i\leq l \mbox{ and } y_i=j\\
    0 & \mbox{otherwise}
    \end{array}
    \right. \;.
\end{equation}
The pseudo-labels are obtained by diffusing the label matrix $\mY$ according to $\adjmatrix$ until convergence (theoretical details in~\cite{zhou2004learning} and practical implementation in~\cite{iscen2019label}):
\begin{equation}
    \mZ = (\identity - \alpha\adjmatrix)^{-1}\mY\;,
\end{equation}
where $\alpha$ is a hyperparameter. The pseudo-label $\hat{y_i}$ of an unlabeled sample $x_i$ is finally defined as:
\begin{equation}
    \hat{y_i} = \text{arg}\,\max_j \vz_{i,j}\;.
\end{equation}
The set of pseudo-labels attributed to the unlabelled samples is noted $\hat{Y}_U$. A weight is associated with each pseudo-label to measure its uncertainty.
Finally, the DNN is fine-tuned for a few epochs to minimize a new weighted loss that considers the pseudo-labels. The new loss is defined as:
\begin{equation}
    L_w(X_S, Y_S, X_U, \hat{Y}_U, \theta) = \sum_{i=1}^l \zeta_{y_i}l_s(\featureextractor(\vx_i, y_i) + \sum_{i=l+1}^n \omega_i\zeta_{\hat{y}_i}l_s(\featureextractor(\vx_i, \hat{y}_i)\;,
\end{equation}
where $\zeta_j$ is the reciprocal of the number of samples assigned to class $j$ (known labels and pseudo labels).

The authors of~\citep{iscen2019label} perform experiments using the CIFAR-10 and mini-Imagenet datasets. Both datasets are split into a training and a test set. In the training set, a number of images are labeled, the others are not.

In the semi-supervised setting, the error is reduced with respect to the fully-supervised setting. Without any surprise, the gain is observed to be decreasing when the number of labeled data points increases. 
To summarize, in this part it was described how to denoise the features of the data samples in three applications, where the graph filtering of either features or labels is considered.
In the features case, it is assumed that the representations of all data samples should live on a manifold, and that the extracted features are noisy versions of these ``ideal'' representations. So, using a graph framework, it is possible to average the representation of the images with the most similar representations obtained with other images. In the second case, images having similar representations are assumed to have similar labels. Thus, the main goal is to smooth the label signal on the similarity graph whose vertices are the images and the edge weights are the similarity between the representations of two images.

\subsection{Graphs as Losses}\label{losses}

Training a DNN for classification often requires to optimize it in order to minimize a cross-entropy loss. As stated in Section~\ref{dnn-cv}, the cross-entropy loss introduces several biases in the learning process. This section gathers several alternative graph-based losses showing interesting properties.

\subsubsection{Smoothness-based loss}\label{losses:smoothness}

In this subsection, a graph-based loss handling the limitations of the very popular cross-entropy loss is presented.
Recall that the cross-entropy loss requires the outputs of the network to converge to label indicator vectors, which comes with several shortcomings described in Section~\ref{dnn-cv}.

Authors have proposed solutions to overcome these drawbacks. For example, in~\citep{hermans2017defense}, the authors use triplets as loss instead of the cross-entropy. In a triplet, the first element is the example to train, the second belongs to the same class and the last to another class. They update the parameters of the DNN so that the first example is closer to the second example than to the last one. Here, a similar loss based on GSP is presented. Contrary to the previous loss, the so-called graph smoothness loss only aims at maximizing the distances between outputs of different classes~\citep{bontonou2019smoothness}. To avoid the DNN converging to a trivial solution that would scatter the outputs far away from each other regardless of their class, the outputs are constrained to remain in a compact subset of the output space.

Consider the problem of training a DNN $f$ to classify examples. The input of the $i$th example is denoted $\vx_i$ and the output of the DNN $\hat{\vy}_i$, such that $\hat{\vy}_i = f(\vx_i)$. Given a distance metric $\|\cdot\|$ and a parameter $\alpha$, a graph $\gG$, whose vertices are examples, is generated. Its edges correspond to the similarity between the outputs of the examples. Thus, the adjacency matrix $\adjmatrix$ of $\gG$ is given by: 
\begin{equation}
    \eadjmatrix_{i, j} \neq 0 \Rightarrow \eadjmatrix_{i, j} = \exp{\left(-\alpha\|f(\vx_i) - f(\vx_j)\|\right)},\forall i, \forall j\;.
\end{equation}
The matrix $\adjmatrix$ is then thresholded in order to only keep the $k$ biggest edges between the vertices. 
Finally, to store the information about a label $c$, a signal on $\gG$ is defined as a binary indicator vector $\vs_c \in \R^{|\sV|}$. Hence, $\evs_{c}[i] = 1$ if and only if $\vx_i$ has $c$ as label.

According to the definition of the smoothness in Section~\ref{gsp:smoothness}, the graph smoothness loss over the graph $\gG$ is given by:
\begin{eqnarray*}
   \mathcal{L}_{\gG} &=& \sum_{\forall c}  {\vs_c}^{\top} \mL \vs_c\\
   &=& \hspace{-1.2cm}\underbrace{\sum_{\genfrac{}{}{0pt}{2}{\vx_i, \vx_{j}, \emW_k[i, j] \neq 0}{ \evs_c[i] \evs_{c}[j] = 0, \forall c}}}_{\text{sum over inputs of distinct classes}}{\hspace{-1cm}\exp{\left(-\alpha\|f(\vx_{i}) - f(\vx_{j})\|\right)}}
   \;.
\end{eqnarray*}

The graph smoothness loss addresses the following drawbacks of the cross-entropy loss:
\begin{itemize}
    \item The dimension of the DNN output is less tightly tied to the number of classes with the proposed loss than with the cross-entropy one.
    \item Keeping only $k$ edges per vertex gives more flexibility to the proposed loss: using a small value of $k$, it is possible to minimize the graph smoothness loss with multiple clusters of points for each class.
    \item The graph-smoothness loss is only interested in relative positioning of outputs with regards to one another, and is therefore built upon the initial distribution yielded by the network.
\end{itemize}

In~\citep{bontonou2019smoothness}, the performance of the graph-smoothness loss is evaluated using three common datasets of image classification. For each dataset, a Resnet-18~\citep{he2016deep} is trained with the cross-entropy loss and another one with the graph smoothness one. To perform classification with the graph smoothness loss, an additional classifier (in the paper a K-nearest neighbors classifier with $K=10$) is also trained on top of the DNN.

The authors show that the graph smoothness loss is able to compete in terms of raw performance with the cross-entropy one. They also perform robustness experiments where they observe that the networks trained with the graph smoothness loss are able to better handle additive noise to the processed inputs.

\subsubsection{Graph knowledge distillation}

To achieve state of the art results, DNNs often rely on a large number of trainable parameters, and considerable computational complexity. This is why there has been a lot of interest in the past few years towards their compression, so that they can be deployed onto embedded systems or in real-time settings. The purpose of this subsection is to analyze a graph-based technique, called Graph Knowledge Distillation (GKD), that allows for an efficient compression of the DNN architecture, while maintaining a high level of accuracy. This framework has been already used with success in the literature in~\citep{liu2019knowledge,lee2019graph,lassance2020deep}. 

Distillation-based approaches aim at distilling knowledge from a pre-trained larger network that is called teacher to a smaller yet to be trained network called student.
For simplicity, assume that both architectures always generate the same number of intermediate representations, even if they do not have the same depth. The loss function of the student network trained with knowledge distillation is defined as:
\begin{equation}
  \mathcal{L} = \mathcal{L}_\text{task} + \lambda_{\text{KD}} \cdot \mathcal{L}_\text{KD}\;,\label{distill_loss}
\end{equation}
where $\mathcal{L}_\text{task}$ is the same loss as the one used to train the teacher (e.g. cross-entropy), $\mathcal{L}_\text{KD}$ is the distillation loss and $\lambda_{\text{KD}}$ is a scaling parameter to control the importance of the distillation with respect to that of the task. 

Given an architecture $A$, a batch of inputs $\mX$ and a subset of layers, the set $\sX'$ contains the intermediate representations of $\mX$ after all considered layers. For each layer, the representations after the layer $\mX' \in \sX'$ are used to define a similarity graph $\gG^A(\mX')$. These graphs contain a vertex for each input in the batch, and the weight of the edge between two vertices is inferred using a measure of similarity between the representations of the associated inputs. Finally, in order to control the importance of outliers, the adjacency matrices of the graphs are normalized ($\adjmatrix = \mD^{-\frac{1}{2}}\adjmatrix\mD^{-\frac{1}{2}})$.

While training the student, the training batch of inputs goes through both the student architecture and the (now fixed) previously trained teacher architecture. The loss to minimize combines the task loss, as expressed in Equation~\ref{distill_loss}, with the following graph knowledge distillation (GKD) loss: 
\begin{equation}
    \mathcal{L}_{\text{KD}} = \sum_{\mX' \in \sX'}{\mathcal{L}_d(\gG^S(\mX'),\gG^T(\mX'))}\;,
\end{equation}
where $\mathcal{L}_d$ is the squared Frobenius norm between the adjacency matrices. The GKD loss measures the discrepancy between the adjacency matrices of teacher and student graphs. In this way, the geometry of the intermediate representations of the student is forced to converge to the one of the teacher. The intuition is that since the teacher network is expected to generalize well to the test, mimicking the geometry of its intermediate representations should allow the student network to generalize better. An equivalent definition of the proposed loss is:
\begin{equation}
    \mathcal{L}_{\text{KD}} = \sum_{\mX' \in \sX'}{\|\adjmatrix^S(\mX')-\adjmatrix^T(\mX')\|_2^2}\;.
\end{equation}

In~\citep{lassance2020deep}, the advantage of using GKD compared with a baseline method (RKD-D~\citep{park2019rkd}) is evaluated using CIFAR-10 and CIFAR-100 datasets~\citep{krizhevsky2009learning}. The obtained results are analyzed on their: \begin{inlinelist} \item accuracies \item decision consistency \item spectral representations \end{inlinelist}.

In terms of accuracy, the authors compare student sized networks trained without distillation (Baseline),
with GKD and RKD-D GKD methods. The comparison shows that RKD-D by itself provides a small gain in error rate with respect to the Baseline approach, while GKD outperforms RKD-D by almost a similar amount. Furthermore, the performance of GKD are shown to be better when using task specific graphs. In other words, by removing the edges between elements of the same class. GKD with task specific graphs provides a gain over GKD of the same magnitude as GKD over RKD-D. Thus, by going from RKD-D to GKD-task specific it was possible obtain a twofold gain over the baseline. 

In terms of decision consistency, the authors evaluate the consistency by taking the trained students and comparing their outputs to the trained teacher's outputs.
It was reported that the ideal scenario that greatly improves the classification performance occurs when the student is 100\% consistent with the teacher's decision on the test set. The experiment shows that the GKD is more consistent with the teacher than the RKD-D.

In terms of spectral representations, it is quite natural to analyze performance from a GSP perspective~\citep{shuman2013emerging}. Considering specific graph signals, the respective smoothness on each of the two graphs (RKD-D and GKD) are compared. Two signals are considered: \begin{inlinelist} \item the label binary indicator signal \item the Fiedler eigenvectors from each intermediate representation in the teacher.
\end{inlinelist} Experiments show that both signals are smoother in the networks trained with GKD. This means that the geometry of the intermediate spaces from GKD are more aligned to those of the teacher. 

\subsubsection{Affinity-based loss}

In this subsection, we present another way of defining the loss of a DNN using graphs~\cite{wang2020affinity}. During training, a DNN is evaluated on a whole batch of data samples before being updated. The idea described in~\cite{wang2020affinity} is to ease the training by exploiting meaningful relationships between these data samples. In short, the idea amounts to add a regularization term to the usual loss of the DNN. While the usual loss is only focused on the task to achieve, the regularization term is more representative of the relationships between the data samples within the DNN.
 
Let us consider that a DNN is trained on a batch of $N$ examples. The input of the ith sample is denoted $\vx_i$. The representation of $\vx_i$ obtained after a given layer of the DNN is denoted $f(\vx_i)$. Depending on the task, $f$ can represent a hidden layer or the output of the DNN.

The affinity between two samples $\vx_i$ and $\vx_j$ within the DNN can be computed with any similarity function $\mathcal{W}$, as 
\begin{equation}
\label{Affinity-adj}
  \eadjmatrix_{i, j} = \mathcal{W} (f(\vx_i), f(\vx_j)). 
\end{equation}
Then, to model the relationships between the examples within the DNN, a graph $\gG$ is defined. Its vertices are the examples. The connection between two examples $i$ and $j$ is weighted by their affinity $ \eadjmatrix_{i, j}$. The adjacency matrix of this affinity graph is denoted by $\adjmatrix$.

Now, the idea in~\cite{wang2020affinity} is to select a meaningful set of connections that we will aim at maximizing during training. For instance, in a classification task, we want to maximize the similarity between the outputs of data samples of the same class. So, the set of meaningful connections amounts to the set of connections between the examples of the same class. Formally, to select the relevant connections, we need to define a target matrix $\boldsymbol{\mathcal{T}} \in \mathbb{R}^{N\times N}$. Given $S$ the set of meaningful pairwise connections, the target $\boldsymbol{\mathcal{T}}$ is defined as

\begin{equation*}
\label{adjacency-matrix}
  \mathcal{T}_{i,j}= \left\{
    \begin{array}{ll}
        1, &  \mathrm{if} \  (i, j) \ \in S,\\
       0, & \mathrm{otherwise}.
    \end{array}
\right.
\end{equation*}

Finally, given $\hat{\adjmatrix}= \mathrm{softmax}(\adjmatrix)$ a matrix-wise softmax operation, the quantity we aim at maximizing is:
\begin{equation*}
    \mathcal{M} = \sum \hat{\adjmatrix} \odot \boldsymbol{\mathcal{T}}.
\end{equation*}

In~\cite{wang2020affinity}, the maximization of $\mathcal{M}$ is expressed as the minimization of a loss $\mathcal{L}_\gG$. A discussion on the choice of the loss form is made in the original paper~\cite{wang2020affinity}.

Finally, the total loss of the DNN is given by 
\begin{equation*}
    \mathcal{L}= \mathcal{L}_{task}+\lambda \mathcal{L}_\gG,
\end{equation*}
where $\mathcal{L}_{task}$ could be the cross-entropy loss, and $\lambda$ is a parameter that takes values in $[0, 1]$. The minimization of the affinity loss $\mathcal{L}_\gG$ places greater emphasis on the connections in $S$ during training.

In~\citep{wang2020affinity}, experiments are made on CIFAR-10, CIFAR-100 and Tiny-ImageNet are summarized. The experiments show that there is an improvement in terms of classification accuracy over a baseline, when using the affinity-based loss. In particular, for datasets with a large number of categories, such as CIFAR-100 (100 classes) and Tini-ImageNet (200 classes), the performance gain is above 1\%.

\subsection{Graphs as Regularizers}\label{regularizers}

In this section, a graph-based technique is used to improve the robustness of DNNs to deviations of the inputs. In other words, the goal is to ensure that a small difference in the input does not lead to a large difference in the output (decision) of the DNNs.

\subsubsection{Bounding smoothness evolution between layers}\label{regularizers:laplacian_nets}

In this subsection, a regularizer that penalizes large deformations of the class boundaries (i.e. distances between the closest elements of distinct classes) throughout the network architecture is presented~\citep{lassance2019robustness}. The regularizer is defined independently of the types of perturbations that is expected to face when the system is deployed.

The regularizer is based on a series of graphs, one for each layer of the DNN. Each graph captures the similarity between the intermediate representations of training examples at that layer.

Given a batch of $b$ inputs $\{\mathbf{x}_1,\dots,\mathbf{x}_b\}$ and a similarity measure $\text{sim}$, the adjacency matrix of the graph generated by intermediate representations at layer $\ell$ is defined as: 
\begin{equation}
    \emAdjacency^\ell_{i,j} = \text{sim}(\vx_i^{\ell+1}, \vx_j^{\ell+1}), \forall 1\leq i,j \leq b.
\end{equation}
In the experiments, the cosine similarity is mainly used.\footnote{It is often the case that the output $\vx^{\ell+1}$ is obtained right after using a ReLU function, that forces all its values to be nonnegative, so that all values in $\adjmatrix^{\ell}$ computed with the cosine similarity are between 0 and 1.} 

To measure the deformation induced by a given layer $\ell$, a regularizer $\delta_\sigma^\ell$ is computed as the difference between label signal smoothness before and after the layer for all labels. More precisely,

\begin{equation}\delta_\sigma^\ell = \sum_c{\left|\sigma^\ell(\mathbf{s}_c) - \sigma^{\ell-1}(\mathbf{s}_c)\right|}\;,\label{eq-reg}\end{equation}
where $\sigma^\ell$ is the smoothness (Section~\ref{gsp:smoothness}) of the binary label indicator vector on the intermediate representation graph from layer $\ell$. These quantities are then used to regularize modifications made by each of the layers during the learning process.

The regularizer penalizes large changes in the smoothness (computed using the Laplacian quadratic form) of the class indicator vectors (viewed as ``graph signals'') between layers. As a result, the margin is kept almost constant across layers, and the deformations of space are controlled at the boundary regions. 
More details can be found in the article~\citep{lassance2020laplacian}.

Since only label signals are considered, only the similarities between examples of distinct classes are of interest. As such, the regularizer only focuses on the boundary, and does not vary if the distance between examples of the same label grows or shrinks.

The experiments show that the regularizer is able to increase robustness to random perturbations and to weak adversarial attacks on the CIFAR-10 dataset.
To validate the generality of the method, the reader is referred to~\citep{lassance2018laplacian,lassance2020laplacian}, where the analysis is extended to two other datasets (CIFAR-100~\citep{krizhevsky2009learning} and Imagenet32x32~\citep{chrabaszcz2017downsampled}).  In the following, the regularizer robustness to some common perturbations is discussed:
\begin{inlinelist}
\item noise~\citep{hendrycks2019robustness,mallat2016understanding}
\item adversarial attacks~\citep{szegedy2014intriguing,goodfellow2014adversarial}
\item implementation defects.
\end{inlinelist}

The robustness of the network to noise is evaluated.
The benchmark from~\citep{hendrycks2019robustness} is used. The benchmark consists in 15 different perturbations, with 5 levels of severity each.  
it is observed that the regularizer is able to increase the relative performance under perturbations by a significant amount (using the mREI measure from~\citep{lassance2020laplacian} an improvement of 0.29 is noticed).

The robustness to adversarial inputs is evaluated, which are specifically built to fool the DNN. The network performance is evaluated by adding the scaled gradient sign (FGSM attack) to the input~\citep{kurakin2017adversarial}, so that a target SNR of 33 is obtained.
The experiments show that the regularizer increases robustness against adversarial attacks as the regularizer improves the network's error rate from 66\% to 49\%. 

The robustness of the architecture to noise on parameters and activations is evaluated. Two types of noises are considered: \begin{inlinelist} \item erasures of the memory (dropout) \item quantization of the weights~\citep{hubara2017quantized}.\end{inlinelist} 

Both experiments show that the regularizer is able to improve the robustness. In the case of the dropout experiment, the regularizer improves the error rate from 55\% to 34\% and in the case of quantization the error rate improves from 76\% to 50\%.

\subsubsection{Peernets: mixing graph and image filters to improve robustness}

In this subsection a second graph-based method which specifically addresses the problem of adversarial attacks~\cite{svoboda2019peernets} is presented. Small perturbations on an input image, often invisible for the human eye, can mislead a DNN and completely distort its prediction. 
The authors of~\cite{svoboda2019peernets} propose to modify the architecture of standard DNNs to make DNNs more robust to adversarial attacks. 

The intuition was that smoothing some fragments of the intermediate representations of the images with a linear combination of fragments from intermediate representations of other similar images would improve the robustness without damaging too much the accuracy. In this way, this should allow one to ``filter away'' (with a graph filter) the adversarial noise. 

The authors of~\cite{svoboda2019peernets} propose to implement this concept as a new layer, called the \emph{Peer Regularization layer}, to intertwine within standard layers of existing DNN architectures. They have been inspired from various domains including Deep learning on graphs, low-dimensional regularization and non-local image filtering. In the following, a detailed description of the Peer Regularization layer is given.

Consider $N$ images (the peers) and their intermediate representations at the output of a given layer. The representation of an image is a $n \times \features$ matrix, where $n$ is the number of pixels and $\features$ is the number of features in each pixel.

The idea of the Peer Regularization layer is to leverage the structure of the feature space of the intermediate representations obtained after a conventional layer (e.g. after a CNN). Each pixel of each intermediate representation is replaced by a smoothed version of itself. The smoothed version is computed from the $K$ most similar pixels among all the pixels belonging to the intermediate representations of all the other images (the peers). More precisely, for each pixel of each intermediate representation, a $K$ nearest neighbor graph is computed. This means that the similarities (for instance, cosine similarities) are computed between the $\features$-dimensional vector representing the pixel and all $\features$-dimensional vectors representing pixels of all the peer images. Then, only the $K$ pixels showing the biggest similarities are kept to compute the smoothed version of the initial pixel.

The $\features$-dimensional vectors of the $K$ pixels are combined using an attention mechanism. Let us denote the $p$th pixel of image $i$ $\vx_p^i$. Its $k$th neighbor is a pixel $q_k$ from an image $j_k$ written $\vx_{q_k}^{j_k}$. The smoothed version of the pixel $\tilde{\vx}_p^i$ is given by the equation:
\begin{equation}
\label{eq:PR_layer}
\tilde{\vx}^{i}_{p} = \sum_{k=1}^K \alpha_{i j_k p q_k} \vx^{j_k}_{q_k}, \hspace{7.5mm} 
\alpha_{i j_k p q_k} = \frac{\mathrm{LeakyReLU}(\exp(a(\vx^i_{p},\vx^{j_k}_{p_k})))
}{
\sum_{k'=1}^K \mathrm{LeakyReLU}(\exp(a( \vx^i_{p},\vx^{j_{k'}}_{p_{k'}} )))
}\,.
\end{equation}
The function $a$ is a fully connected layer whose input is a $2\features$-dimensional vector and outputs a scalar. The $\alpha$ represents the relative importance of each neighboring pixel.

In the original paper, the authors report experiments on CIFAR-10 with a Resnet-32 containing intertwined Peer Regularization layers (PR-Resnet-32). The test images are disrupted with three white-box adversarial attacks (gradient descent, fast-gradient sign method, projected gradient descent) of varying amplitude (measured by the variable $\epsilon$). 
It was shown that PR-Resnet-32 is more robust under these adversarial attacks than the standard Resnet-32 (more details in the original paper~\cite{svoboda2019peernets}). 

While this kind of methods is not perfect yet, it shows an interesting example of how graph methods can help to handle a Deep Learning problem. Here, the graph enables to take advantage of the relationships between data at any point within a DNN, and to use this knowledge to make it more robust to small perturbations. Note however, that it requires additional training time and arbitrary choices about the training images to keep during testing. More information on the implementation can be found in the original paper.  

\section{Conclusion}
\label{conclusion}
DNNs achieve state of the art performance in many machine learning tasks. Unfortunately, their practical uses in real-world applications are still limited due to some major limitations.
In this chapter, we have presented an overview of graph-based tools that aim at solving some of these limitations. Namely, \begin{inlinelist} \item interpreting the intermediate representations of DNNs \item train models with few data \item learning meaningful embeddings for classification  \item improving the robustness of DNNs \item compressing the architectures \end{inlinelist}. 

Approaching these limitations with graphs enables to view them from a new angle. In many methods, the graphs are used to model the intermediate spaces of the DNNs.
In general, their vertices represent distinct data samples and their edges depend on the similarity of the intermediate representations of the data samples within the DNNs. Such graphs are called \emph{latent geometry graphs}. Given the ubiquity of graphs in representing relations between data samples, it is quite natural to imagine that GSP, a branch of the signal processing field that handles data supported on graphs, is a valuable path to address these challenges.  

This area of research is in its infancy. We hope that such an overview will inspire researchers to work in this line of research that we believe could help improving the applications of deep learning that we see in day-to-day life.


\bibliography{references}
\bibliographystyle{natbib}

\end{document}